# Fair Marriage Principle and Initialization Map for the EM Algorithm

Chenguang Lu

*Abstract*: The popular convergence theory of the EM algorithm explains that the observed incomplete data log-likelihood $L$ and the complete data log-likelihood $Q$ are positively correlated, and we can maximize $L$ by maximizing $Q$. The Deterministic Annealing EM (DAEM) algorithm was hence proposed for avoiding locally maximal $Q$. This paper provides different conclusions: 1) The popular convergence theory is wrong; 2) The locally maximal $Q$ can affect the convergent speed, but cannot block the global convergence; 3) Like marriage competition, unfair competition between two components may vastly decrease the globally convergent speed; 4) Local convergence exists because the sample is too small, and unfair competition exists; 5) An improved EM algorithm, called the Channel Matching (CM) EM algorithm, can accelerate the global convergence. This paper provides an initialization map with two means as two axes for the example of a binary Gaussian mixture studied by the authors of DAEM algorithm. This map can tell how fast the convergent speeds are for different initial means and why points in some areas are not suitable as initial points. A two-dimensional example indicates that the big sample or the fair initialization can avoid global convergence. For more complicated mixture models, we need further study to convert the fair marriage principle to specific methods for the initializations.

## 1. Introduction

The EM algorithm [1] is sensitive to initial parameters so that little different parameters may vastly change the speed, even the result of convergence [2,3], especially when the overlap of components is serious [4]. Hence, most improvements focus on the initial parameters, such as in the Deterministic Annealing EM (DAEM) algorithm proposed by Ueda and Nakano [2], the Split and Merge EM (SMEM) algorithm proposed by Ueda et al. [5], the Competitive EM (CEM) algorithm [6], the Random swap EM algorithm [7], the cross-entropy method with EM algorithm [8], and robust EM algorithms in [9]. Some improved EM algorithms can reduce 30%-60% iterations that the EM algorithm needs, as reported in [10,11,12].

The popular convergence theory of the EM algorithm can be found in the articles of Dempster *et al*. [1], Wu [13], and Little *et al*. [14]. They explain that by repeating the E-step and the M-step, we can make the log-likelihood $L=L(\mathbf{X}|\theta)=\log P(\mathbf{X}|\theta)$ converge to its maximum, where $\mathbf{X}$ denotes the observed incomplete data, because

**Affirmation I**: The incomplete data log-likelihood $L$ increases with the complete data log-likelihood $Q=\log P(\mathbf{X}, \mathbf{Y}|\theta)$, where $\mathbf{Y}$ denotes unobserved data; we can maximize $L$ by maximizing $Q$.

**Affirmation II:** $Q$ is increasing in the M-step and is non-decreasing in the E-step, or $Q$ is increasing in every iteration.

However, it has never been well proved that $Q$ is non-decreasing in the E-step, or $Q$ is increasing in every iteration.

After analyzing the EM algorithm by semantic information methods [17,18] and programming practices, the author of this paper found that the EM algorithm is better than many researchers expect because the EM algorithm can get away from locally maximal $Q$ to achieve globally maximal $L$ in most cases; some local convergence comes only because of misunderstanding (see Section 3.1). He also found that the popular convergence theory of the EM algorithm is worse than most researchers believe because the above two affirmations are wrong.

In the author's previous articles [15,16], he introduced the CM-EM algorithm and new convergence proof. The primary purposes of this paper are
1)  to explain the convergence difficulties by the marriage competition interpretation, and

2) to provide an initialization map for binary Gaussian mixture to show the convergence difficulties of using different initial means.

This map must be useful for the initialization of the EM or an improved EM algorithm for binary Gaussian mixture models and be helpful for other mixture models.

## 2. The Problems the Popular Convergence Proof of the EM algorithm

### 2.1 The EM algorithm and the CM-EM algorithm

**Definition 1**. Let $x \in U=(x_1, x_2,...x_m)$ be an instance with prior distribution $P(x)$, which is also the sampling distribution for the incomplete data.

**Definition 2**. A sample **D** consists of $N$ examples, e.g., **D**=$\{(x(t); y(t)), t=1, 2, ..., N; x(t)\in U; y(t)\in V\}$. A conditional sample is **D**$_j$=$\{(x(t); y_j), t=1, 2, ..., N_j; x(t)\in U\}$, where $N_j$ is the number of examples with $y_j$.

**Definition 3**. Let $\theta$ be a predictive model. For each $y_j$, there is a sub-model $\theta_j$. We define $P(x|y_{\theta j})=P(x|y_j,\theta)$, which is a predictive distribution and also the likelihood function of $\theta_j$.

Hence the predicted distribution of $x$ is

$$P_\theta(x) = \sum_j P(y_j)P(x|y_{\theta j}). \qquad (1)$$

To solve a mixture model is to find $P_\theta(x)$ that maximize the observed incomplete data log-likelihood or minimize the relative cross-entropy:

$$H(P||P_\theta) = \sum_i P(x_i)\log\frac{P(x_i)}{P_\theta(x_i)} = H_\theta(X) - H(X). \qquad (2)$$

Steps in the EM algorithm for Gaussian Mixtures are:

- **E-step**: Write the conditional probability functions for the Shannon channel [19]:

$$P(y_j|x)=P(y_{\theta j}|x) = P(y_j)P(x|y_{\theta j})/P_\theta(x),\ j=1,2,...,n;$$
$$P_\theta(x) = \sum_j P(y_j)P(x|y_{\theta j}). \qquad (3)$$

- **M1-step**: Calculate new mixture proportions:

$$P^{+1}(y_j) = \sum_i P(x_i)P(y_j|x_i) = \sum_i P(x_i)P(x_i|y_{\theta j})P(y_j)/P_\theta(x_i),\ j=1,2,...,n. \quad (4)$$

- **M2-step**: Calculate new parameters:

$$\mu_j^{+1} = \sum_i P(x_i|y_{\theta j}^{+1})x_i,\ j=1,2,...,n;$$
$$\sigma_j^{+1} = \{\sum_i P(x_i|y_{\theta j}^{+1})[x_i - \mu_j^{+1}]^2\}^{0.5},\ j=1,2,...,n. \qquad (5)$$

For a sample with size $N$: $\{x(t)|t=1,2,...N, x(t)\in U\}$, we have

$$\mu_j^{+1} = \frac{1}{N}\sum_{t=1}^N P(x(t)|y_{\theta j}^{+1})x(t),\ j=1,2,...,n;$$
$$\sigma_j^{+1} = \{\frac{1}{N}\sum_{t=1}^N P(x(t)|y_{\theta j}^{+1})[x(t) - \mu_j^{+1}]^2\}^{0.5},\ j=1,2,...,n. \qquad (6)$$

If $U$ is two-dimensional, we need to calculate the correlation coefficients.

The author proposed the Channel Matching (CM) EM algorithm [15,16]. It includes two steps:
- To repeat the E-step and the M1-step until $P(y_j^{+1})=P(y_j)$ for all $j$.
- To execute M2-step one time.

We call the CM-EM algorithm as the E3M algorithm if we only repeat the E-step and the M1-step three times.

### 2.2. A Counterexample with Q>Q* against the two affirmations

To prove that the above Affirmation I and Affirmation II are the two theoretical mistakes in the popular convergence theory of the EM algorithm, let us see a counterexample.

A true mixing proportion distribution $P^*(y)$ and a true conditional probability distribution $P^*(x|y)$ ascertain the joint probability distribution $P^*(x, y)= P^*(y)P^*(x|y)$. The corresponding joint entropy is

$$H^*(X,Y) = -\sum_j \sum_i P^*(x_i, y_j) \log P^*(x_i, y_j) = -Q^*/N. \tag{7}$$

Now we show that $Q$ may be greater than $Q^*$.

**Example 1.** $U=\{1, 2, 3, \ldots, 150\}$; a true model is $(\mu_1^*, \mu_2^*, \sigma_1^*, \sigma_2^*, P^*(y_1))=(65, 95, 15, 15, 0.5)$. Suppose that the guessed ratios and parameters are $P(y_1)=0.5$, $\mu_1=\mu_1^*$, $\mu_2=\mu_2^*$, and $\sigma_1=\sigma_2=\sigma$. Fig. 1 shows that $Q$ and $L$ change with $\sigma$. The source file in Python 3.6 for Examples 1-3 can be found in Appendix I.

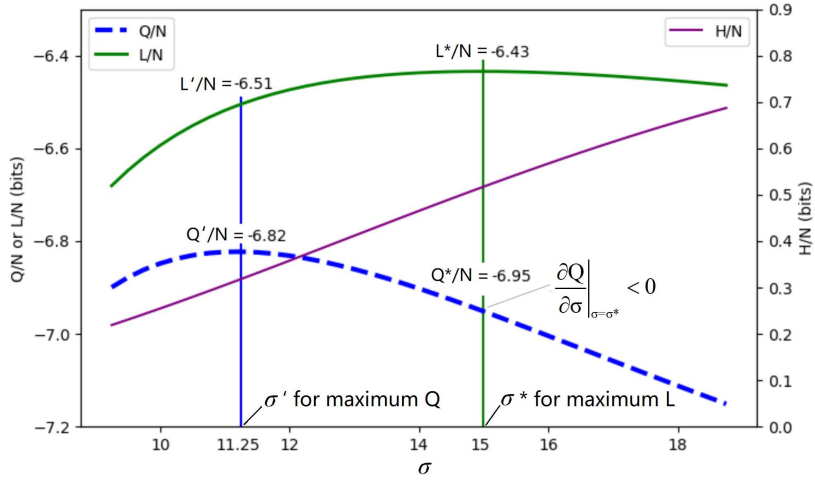

**Fig. 1**. $Q$ and $L$ change with $\sigma$ against the two affirmations.

In this example, when $\sigma$ changes from $\sigma'=11.25$ to $\sigma^*=15$, $Q$ decreases from its global maximum $Q'$ =-6.82N bits to $Q^*$=-6.95N bits; whereas $L$ increases from $L(\sigma')$=-6.51N bits to its global maximum $L(\sigma^*)$=-6.43N bits. Therefore, the Affirmation I in the popular convergence proof of the EM algorithm is wrong. We can use the EM algorithm to solve the above example with an initial $\sigma$=11.25. Fig. 2 (c) shows that $Q$ changes with $H(P||P_\theta)$ in the EM algorithm. The sample size is 50000.

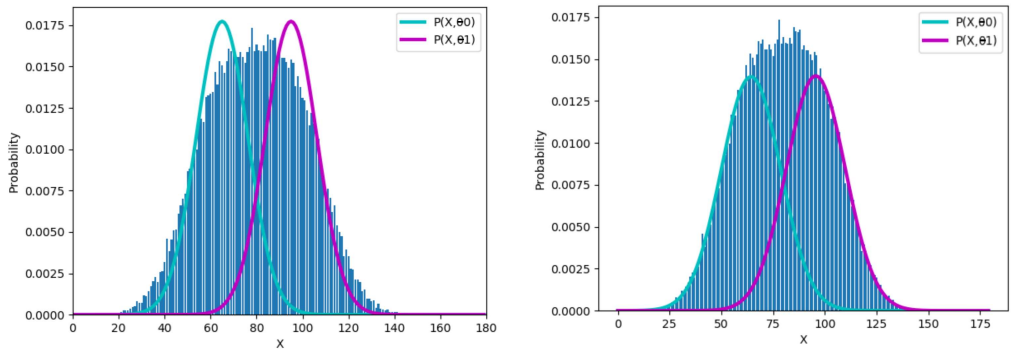

(a) The initial state with larger $Q$= -6.82N bits.  (b) The valid convergence with smaller $Q^*$= -6.95N bits.

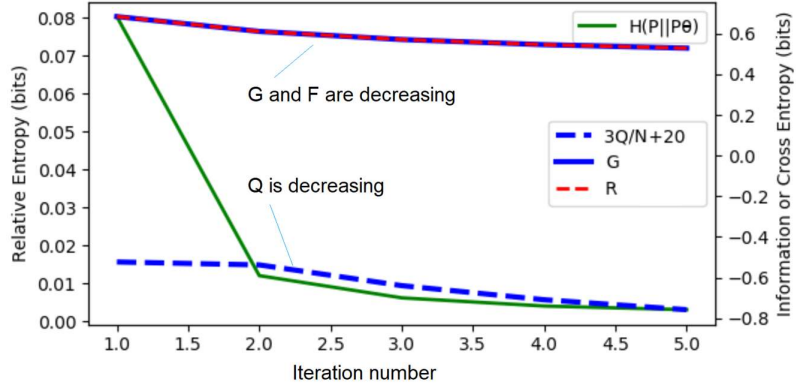

(c) $Q$ and $F$ decrease as $H(P||P_\theta)$ decreases or as $L=L(\mathbf{X}|\theta)$ increases.

**Fig. 2**. A counterexample against the two affirmations about the EM algorithm.

Fig. 2 indicates that the E-step can decrease $Q$. When $Q>Q^*$, it is because the E-step can reduce $Q$ that $Q$ may converge to $Q^*$. Now we can find that the EM algorithm can converge in some cases not because the E-step is non-decreasing, but because the relative entropy $H(P||P_\theta)$ is decreasing. That means that, in the popular convergence proof of the EM algorithm, the mistake in Affirmation II covers up the error in Affirmation I.

### 2.3. The Mathematical analysis about that Q may be Greater than Q*

Further, the author found that for any true model parameter set $\theta^*$ and the corresponding $Q^*=Q(\theta^*)=Q(\mu_1^*, \mu_1^*, \sigma_1^*, \sigma_2^*, P^*(y_1))$, we can always find a ratio $r$ between 0.5 and 1, such as $r=0.75$, so that $Q(\theta')= Q(\mu_1^*, \mu_1^*, r\sigma_1^*, r\sigma_2^*, P^*(y_1))>Q(\theta^*)$. To prove this conclusion, we need to prove $\left.\frac{\partial Q}{\partial \sigma_j}\right|_{\sigma_j=\sigma_j^*} < 0$. This proof is not easy; nevertheless, we can prove that this conclusion is tenable for some symmetrical mixture models. In Example 1, we further suppose that $P(y_1)=P^*(y_1)=0.5$, $\mu_1=\mu_1^*$, $\mu_2=\mu_2^*$, and $\sigma_1=\sigma_2=\sigma$. Hence, $Q$, $L$, and $H$ only change with $\sigma$. We have $Q=Q(\sigma)$, $L=L(\sigma)$, and $H=H(\sigma)$. $L(\sigma^*)$ is the maximum of $L$. Now, we can prove $\left.\frac{dQ}{d\sigma}\right|_{\sigma=\sigma^*} < 0$.

**Proof**: According to Eq. (12) for the E-step,

$$P(y_1|x) = \frac{P(y_1)e^{-(X-\mu_1)^2/(2\sigma^2)}}{P(y_1)e^{-(X-\mu_1)^2/(2\sigma^2)} + P(y_2)e^{-(X-\mu_2)^2/(2\sigma^2)}} = \frac{1}{1+e^{b/\sigma^2}}. \quad (8)$$

where $b=[(\mu_2-\mu_1)x-(\mu_1^2-\mu_2^2)]/2$. We can prove $dH/d\sigma>0$ (see Appendix I for the detailed proof).

Since $dL/d\sigma=0$ as $\sigma=\sigma^*$ is the condition that maximizes $L$, and $Q=L-H$, we have

$$\left.\frac{dQ}{d\sigma}\right|_{\sigma=\sigma^*} = \left.\frac{dL}{d\sigma}\right|_{\sigma=\sigma^*} - \left.\frac{dH}{d\sigma}\right|_{\sigma=\sigma^*} = 0 - \left.\frac{dH}{d\sigma}\right|_{\sigma=\sigma^*} < 0. \quad (9)$$

**Q.E.D.**

The author has tested many true model parameters. For every $\theta^*$, we can always find $\theta'$ so that $Q(\theta')$ is greater than $Q(\theta^*)$ by replacing $\sigma_j^*$ with $\sigma_j'=r\sigma_j^*$ ($r$ is about 0.75).

### 2.4. The Shannon channel P(Y|X) from the E-step for the expectation is abnormal

Shannon [19] calls the conditional probability matrix $P(Y|X)$ as the channel. A Shannon's channel consists of a group of transition function $P(y_j|x)$, where $y_j$ is a constant, and $x$ is a variable. $P(y|x)$ that E-step uses for the expectation is Shannon's channel. A good Shannon's channel $P(y|x)$ should makes

$$P(y_j) = \sum_i P(x_i)P(y_j|x_i) = \sum_i P(x_i|y)P(y_j) = P(y_j), j=1,2,...,n. \quad (10)$$

However, in the E-step, for given $P(x)$, $P(y)$ and $\theta$, we have a new $P(y)$ denoted by $P^{+1}(y)$:

$$P^{+1}(y_j) = \sum_i P(x_i)P(y_j|x_i) = \sum_i P(x_i)P(x_i|y_{\theta j})P(y_j)/P_\theta(x_i). \quad (11)$$

Generally, there is $P^{+1}(y) \neq P(y)$. Therefore, the Shannon's channel from the E-step is abnormal. Even if $P^{+1}(y) \neq P(y)$ is not a mistake, at least it is improper. To provide a proper Shannon's channel, we need to find the unique $P(y)$ that matches $P(x)$ and $\theta$ so that $P^{+1}(y)=P(y)$. That is why the CM-EM algorithm repeats Equations (3) and (4).

## 3. The Marriage Competition Interpretation of Mixture Models

### 3.1. Why does the typical local convergence happen?

It is according to the popular convergence theory of the EM algorithm that Ueda and Nakano [2] proposed the Deterministic Annealing EM (DAEM) algorithm. They conclude that when some initial parameters result in locally maximal $Q$, local or invalid convergence is inevitable; we may use the deterministic annealing method, which increases standard deviations, to avoid locally maximal $Q$. Example 2 is provided in [2] and also verified by Marin *et al.* [3].

**Example 2.** A mixture model has two Gaussian components. The true model is ($\mu_1^*$, $\mu_2^*$, $\sigma_1^*$, $\sigma_2^*$, $P^*(y_1)$)= (0, 2.5, 1, 1, 0.7) (see Fig. 3 in [3]). Marin *et al.* show that among the five initial points on the $\mu_1$-$\mu_2$ plane, using the EM algorithm, only two points validly converge to ($\mu_1^*$, $\mu_2^*$); the other three points invalidly converge to a point near ($\mu_1$, $\mu_2$)=(1.5, -0.5).

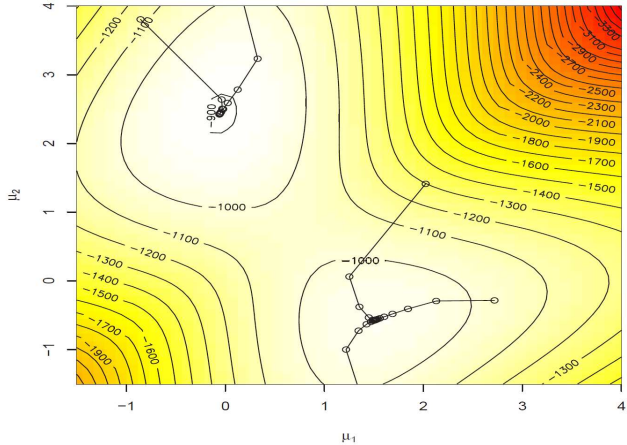

**Fig. 3** Trajectories of five runs of the EM algorithm on the log-likelihood surface (thanks to the authors of [3] for the kind permission).

In [3] and Fig. 3, contour lines represent the log-likelihood $L$. In [2], contour lines represent $-F$ for annealing parameter $\beta=1$; $-F$ is similar to $L$. Both authors believe that point (1.5, -0.5) has locally maximal $Q$ and also locally maximal $L$.

Later, to avoid the local convergence and boundary convergence, Ueda et al. used the Split and Merge EM (SMEM) algorithm [5]; Marin *et al.* proposed the Population Monte Carlo (PMC) algorithm [3].

After inspecting the behaviors of the EM algorithm for Example 2, with different sample sizes and different initial points on the $\mu_1$-$\mu_2$ plane, the author of this paper reached different conclusions. Fig. 4 is used to explain these conclusions. The ($\mu_1$, $\mu_2$) in Fig. 3 becomes ($10\mu_1+100$, $10\mu_2+100$) in Fig. 4.

These conclusions are related to

- **Symmetry:**
  a) If model ($\mu_1$, $\mu_2$, $\sigma_1$, $\sigma_2$, $P(y_1)$) = (100, 125, 10, 10, 0.7) produces sampling distribution $P(x)$, then model ($\mu_1$, $\mu_2$, $\sigma_1$, $\sigma_2$, $P(y_1)$) = (125, 100, 10, 10, 0.3) also produces the same $P(x)$ (see Fig. 5 (a)). Both models can be considered as true models. In other words, there are two points $a_1$ and $a_2$ in Fig. 4, where $L$ is global maximum, denoted by $L^*$; there are also two points $b_1$ and $b_2$, where $Q$ is local maximum, denoted by $Q'$. The two areas divided by the 45° line (red or solid line) are axisymmetric, with the 45° line as the symmetry axis. This line is like a deep ditch with smaller $Q$ and $L$; the EM algorithm cannot cross this deep ditch.

If the initial point is in this deep ditch, the EM algorithm must invalidly converge to point $f$, which is the shallowest point of the deep ditch.

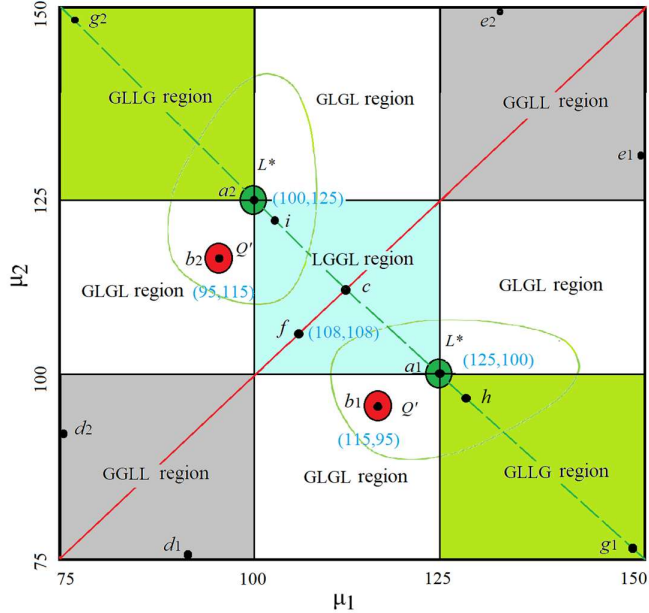

Fig. 4. Symmetrical $\mu_1$-$\mu_2$ plane and the four kinds of regions

b) In Fig. 3 above and Fig. 2 in [10], the contour lines are asymmetrical because the authors [2,3] always use $P^*(y_1)=0.7$ for both sides. However, for the lower side, we should use $P^*(y_2)=0.7$ to draw the contour lines because when the EM algorithm convergences to $b_1$ or $a_1$, $P(y_2)$ is close to 0.7 rather than $P(y_1)$. The authors of [2,3] wrongly think that every convergence to points below the 45° line is local convergence, and every convergence to points above the 45° line is global convergence. Ueda and Nakano [2] provide two bad initial points in the lower side, with which the EM algorithm converges to $b_1$, whereas the DAEM algorithm converges to $a_2$. In this example, the DAEM algorithm makes efforts to cross the deep ditch (the 45° line). However, we need not seek far and neglect what lies close at hand. The $a_1$ is as good as $a_2$. Because the convergence to $a_1$ looks the same as the convergence to $a_2$ (see Fig. 5(b)), someone might mistake $a_1$ for $a_2$.

c) $P^*(y_1) \neq P^*(y_2)$ is the cause of the existence of $b_1$, $b_2$, and $f$. When $P^*(y_1)$ changes from 0.7 to 0.5, points $b_1$, $b_2$, and $f$ disappear or move to $a_1$, $a_2$, and $c$ respectively. Then Fig. 4 becomes centrosymmetric.

- **Local convergence and global convergence**: Local convergence happens only when sample sizes are too small, such as $N=1000$, and the relative positions of ($\mu_1$, $\mu_2$) and ($\mu_1^*$, $\mu_2^*$) are not good. Other points between $b_1$ and (120, 98) or between $b_2$ and (98, 120) are also possibly points of local convergence, which means that the EM algorithm is easily stuck due to fewer sample points. If samples are big enough, such as $N=50000$, with any initial points, the EM algorithm can let $L$ converge to $L^*$. Fig. 5 shows a run with a very bad initial point $d_2=(80,95)$.

- **Relative positions and convergence difficulties**: There are 4!=24 permutations of $\mu_1$, $\mu_2$, $\mu_1^*$, and $\mu_2^*$. If some symmetrical permutations are regarded as the same, all permutations (or nine regions in Fig. 4) can be classified into four kinds represented by four labels: GLLG, LGGL, GLGL, and GGLL. If we put the initial ($\mu_1$, $\mu_2$) in these regions, convergence difficulties increase in turn. When $P(y_1)=P(y_2)=0.5$, Figure 4 is centrosymmetric; all regions with the same label are equally easy for the global convergence. If $P(y_1) \neq P(y_2)$, two regions with the same label are unequally difficult for global convergence. In Fig. 7, iteration numbers that are put in some typical locations represent the convergence difficulties of those locations. In short, (1) points close to the 135° line in GLLG or LGGL regions, such as $g_1$, $g_2$, $h$, and $i$, are best points; (2) points close to the 45° line in GGLL or LGGL regions are worse points; (3) points in or behind a $Q$-larger area, such as areas close to $b_1$, $b_2$, $d_1$, and $d_2$, are worst points.

- **About the annealing operation:** When samples are big enough, it is not necessary to increase $\sigma_1$ and $\sigma_2$, such as using $\sigma_1 = \sigma_2 = 20$, which means that the annealing operation is used. When samples are small, adjusting the relative position or initial $(\mu_1, \mu_2)$ is more effective than adjusting standard deviations.

- **$Q$ changes with $L$**: The EM algorithm can pass through $Q$-larger and $Q$-smaller areas to reach $L^*$ (see Fig. 5); $Q$ and $L$ are not positively correlated. Fig. 5 (a) shows that the two components that result in maximal $Q$. Fig. 5 (b) shows the two components that result in maximal $L$. Fig. 5 (c) shows the iteration process of the EM algorithm for Example 2.

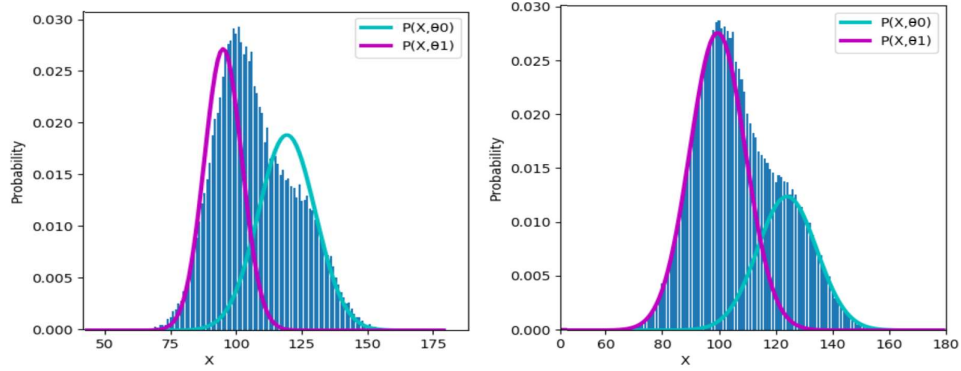

(a) Two components with $(\mu_1, \mu_2)=b_1$ or $b_2$.     (b) Two components with $(\mu_1, \mu_2)=a_1$ or $a_2$.

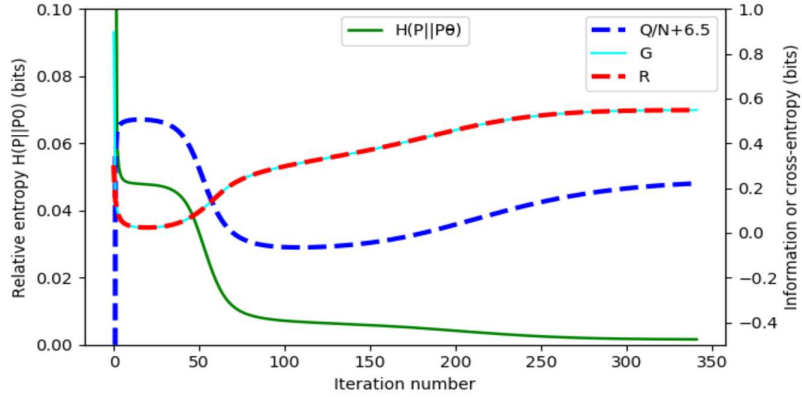

(c)   $Q$ and $H(P||P_\theta)$ changes in iterations (initialization: $(\mu_1, \mu_2, \sigma_1, \sigma_2, P(y_1)) = (80, 95, 5, 5, 0.5)$).

**Fig. 5**. The EM algorithm for Example 2 ($N$=50000) against the popular convergence theory.

In Fig. 5 (c), the initial $\sigma_1$ and $\sigma_2$ are 5 so that $Q'$ is globally maximal. Although $Q'$ is the globally maximal $Q$, it cannot stop $(\mu_1, \mu_2)$ to reach $a_2$ with smaller $Q^*$. The popular convergence theory of the EM algorithm is falsified again.

### 3.2. Explaining convergence difficulties using marriage competition

Why are some initial points on the $\mu_1$-$\mu_2$ plane easy or difficult to converge to $a_1$ and $a_2$ with $L^*$? We use a metaphor to explain the reason. Imagine that two ladies (denoted by L and **L**) and two gentlemen (denoted by G and **G**) live along a road with relative positions like these: "__G__G__L__L__", "__G__L__G__L__", and "__G_____LG__L_". The valid or global convergence means that one G marries with one L. Because of the unfair competition, the left G is not easy to attract the left L, which is also attracted by the right G. If they have cars so that distances are not obstructions, then the valid convergence is a little easy. An L is like a sub-sample, and a G is like a component. The Deterministic Annealing method seemly provides cars to enlarge two gentlemen's scopes of activities. However, it is more effective to adjust the relative positions of two gentlemen. Assume that distances between any two adjacent people are equal. Their relative position can be simply represented by four labels: GLLG, LGGL, GLGL, and GGLL, which reflect the convergence difficulties of different relative positions.

Distances between them also affect convergence difficulties. For example, "__G_____LG__L_" is more difficult for the global convergence than "__G__L__GL__L_" because LG in the former are not easy to separate. If we consider that mixture proportions are different, and a sub-sample with a larger mixture proportion is like a more attractive lady, which is denoted by **L**, then "__G____LG__L_" is more difficult than "_G___LG__L_" for the global convergence. The most difficult relative position is "_G_G___L____L_" (such as $d_1$ and $d_2$ in Fig. 4), which will become "_G_LG__L_", not only because the competition is unfair, but also because it is hard for the right G to give up **L**.

We call the 135° (green or dashed) line as **the fair competition line**, call the 45° (red or solid) line as **the absolute equality line**, and call the area with locally maximal $Q$ as **the hard separation area**. For the better initial means of components, we should follow three principles:

- Let the initial point ($\mu_1, \mu_2$) close to **the fair competition line.**
- Let the initial point ($\mu_1, \mu_2$) apart from **the absolute equality line**, which may result in "all gentlemen marry with all ladies".
- Do not set the initial point ($\mu_1, \mu_2$) in or behind **the hard separation areas**, such as areas around $b_1$ and $b_2$.

We call the above three principles together as **the fair marriage principle**. When two sub-samples have the same means and different standard deviations, perhaps with different correlation coefficients, the competition between two components is probably also unfair, such as in Example 3 (see Section 5.2). In these cases, the fair marriage principle is helpful, but not sufficient. If three components and three sub-samples have relative positions like this: "GGLGLL" or "GGGLLL", for finite-size samples, the left G may be alone, and boundary convergence (a mixing proportion approaches 0) will happen. For avoiding the boundary convergence of more complicated mixture models, the fair marriage principle is also useful, but not sufficient. We need to enrich this principle and convert the enriched principle into specific methods.

The author's experiments indicate that, in most cases, it is better that initial mixture proportions are equal, and initial standard deviations are equal and not too small. Too small standard deviations easily result in boundary convergence for both the EM and CM-EM algorithms.

## 4. Results: three examples showing the CM-EM algorithm's performance

### 4.1. More Experiments about Example 2

The author has used different relative positions to compare the iteration numbers of the EM and E3M (CM-EM with $t=3$) algorithms for Example 2 (see Fig. 6). The true model is ($\mu_1^*, \mu_2^*, \sigma_1^*, \sigma_2^*, P^*(y_1)$) =(100, 125, 10, 10, 0.7). The initialization is ($\mu_1, \mu_2, \sigma_1, \sigma_2, P(y_1)$)=( $\mu_1, \mu_2,$ 7, 7, 0.5). The sample size $N$ is 50000. Two numbers 289/191 at the lower-left corner (($\mu_1, \mu_2$)=(80, 81)) means that using an initial point there, the EM algorithm needs 289 iterations, and the E3M algorithm needs 191 iterations. The other numbers are in like manner. For every initial point, each algorithm repeated three times; the middle iteration number was selected.

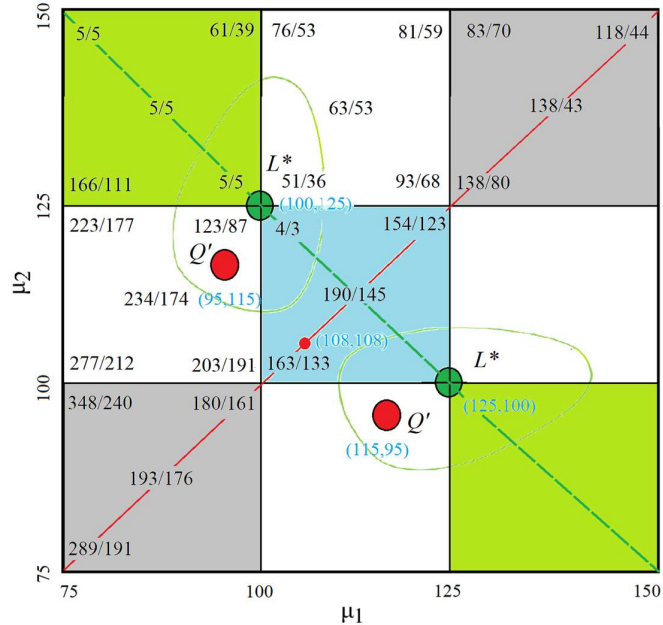

**Fig. 6.** Iteration numbers change with different initial means for Example 2.

The stop condition was that $|\max(\mu_1,\mu_2)-\max(\mu_1{}^*,\mu_2{}^*)|<1$, $|\min(\mu_1,\mu_2)-\min(\mu_1{}^*,\mu_2{}^*)|<1$, $|\sigma_1-\sigma_1{}^*|<1$, $|\sigma_2-\sigma_2{}^*|)<1$, $|\max(P(y_1),P(y_2))-\max(P^*(y_1),P^*(y_2))|<0.033$, and $H(P||P_\theta)<0.005$. Note that in practices, we may only use $H(P||P_\theta)<0.005$ as the stop condition without considering the symmetry. We can also put iteration numbers on the lower-right side according to the axisymmetry. For the iteration numbers at **the absolute equality line**, $(\mu_1,\mu_2)=(\mu_1,\mu_1+1)$ was used as the initial means.

The average iteration numbers are 136.7 (with EM algorithm) and 90.4 (with E3M algorithm). The E3M algorithm needs 74% of iterations that EM algorithm needs. When sample sizes are not big enough, the convergence points are very uncertain. The author examined Example 2 with the EM algorithm and the E3M algorithm, using different sample sizes and initialization. The iteration numbers are shown in Table 1. The stop condition for $N=1000$ is that $|\mu_i-\mu_i{}^*|<1$ and $|\sigma_i-\sigma_i{}^*|<1$, $j=1,2$. The stop condition for $N=100$ is that $|\mu_i-\mu_i{}^*|<3$ and $|\sigma_i-\sigma_i{}^*|<3$, $|P(y_j)-P^*(y_j)|<0.1$, $j=1,2$; $H(P||P_\theta)<0.7$ bit. The samples are produced by a random function so that any two samples in different runs are different.

**Table 1**. Comparison between the CM-EM algorithm and the EM algorithm

|  | N | Initial ($\mu_1, \mu_2$) | Initial $\sigma_1=\sigma_2$ | Time of invalid convergence | Time of IN>500 | Time of fast convergence |
|---|---|---|---|---|---|---|
| EM | 1000 | (95,115) | 10 | 2 | 7 | 5 (IN<100) |
| EM | 1000 | (95,115) | 20 | 1 | 9 | 5 (IN<100) |
| EM | 1000 | (95,115) | 5 | 1 | 9 | 6 (IN<100) |
| CM-EM | 1000 | (95,115) | 10 | 1 | 5 | 10 (IN<100) |
| CM-EM | 1000 | (95,115) | 5 | 0 | 4 | 12 (IN<100) |
| EM | 100 | (80,145) | 10 | 3 | 3 | 16 (IN<4) |
| CM-EM | 100 | (80,145) | 10 | 1 | 1 | 19 (IN<4) |
| CM-EM | 100 | (80,145) | 5 | 2 | 2 | 18 (IN<4) |

*20 runs were used for every initial condition; IN means iteration number.

When the sample size is 1000, although the locally maximal $Q$ is obvious as $\sigma=5$, the EM algorithm performs similarly for $\sigma=5$ and $\sigma=20$. When the sample size becomes 100, and the initial means are fair, both algorithms have less invalid convergence and faster convergent speeds than before. Under every condition, the CM-EM algorithm performs better than the EM algorithm.

Now we can use the CM-EM algorithm and the fair marriage principle to improve the algorithm for Fig. 5 (c). After the first iteration of the E3M algorithm, one mixture proportion becomes a tiny value (0.009), so we can easily find this blocked component. Then we can change the mean of this blocked component to another side of the sample (e.g., change GGLL to GLLG) to accelerate the convergence.

### *4.2. A two-dimensional example for testing the convergence reliability*

Example 3 (see Fig. 7) is used to test whether the CM-EM algorithm can avoid local convergence in two-dimensional instance spaces when the overlap is severe.

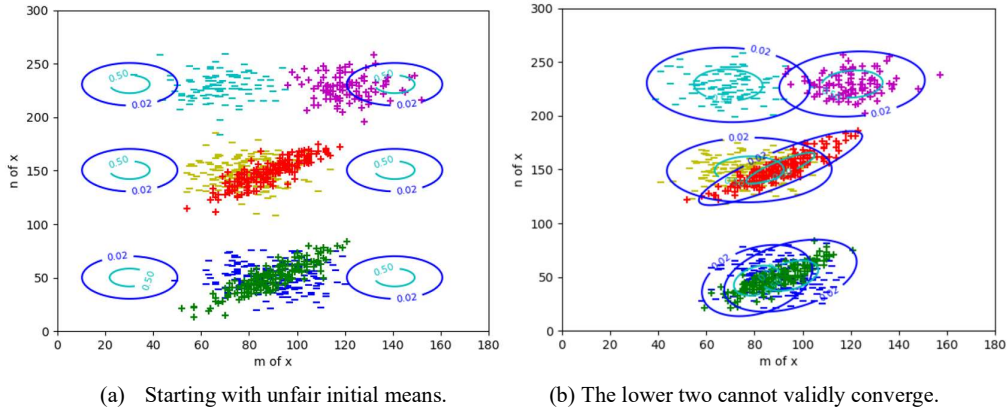

(a) Starting with unfair initial means.     (b) The lower two cannot validly converge.

**Fig. 7.** Example 3 with improper initial parameters.

**Example 3.** Three pairs of samples were used to test whether the CM-EM algorithm could distinguish two components well in every pair (see Fig. 7). The sample size is 1000. The convergence condition is that the horizontal distance between the two centers of the lower two components is smaller than 1 (the best is 0). The True and initial model parameters for Figures 1,2,5-9 can be found in Appendix I.

The upper two pairs could rapidly validly converge. Only the lower pair could not work well because the overlap was severe. After the sample size was changed from 1000 to 50000, the iteration could validly converge after 31 iterations.

The above invalid convergence happened because the competition was unfair since the right component was closer to the two sub-samples than the left one in Fig. 7 (a) so that the hard separation happened.

After the initial means of lower two components were changed for fair competition without absolute equality, as shown in Fig. 8 (a), the iteration could validly converge, as shown in Fig. 8 (b).

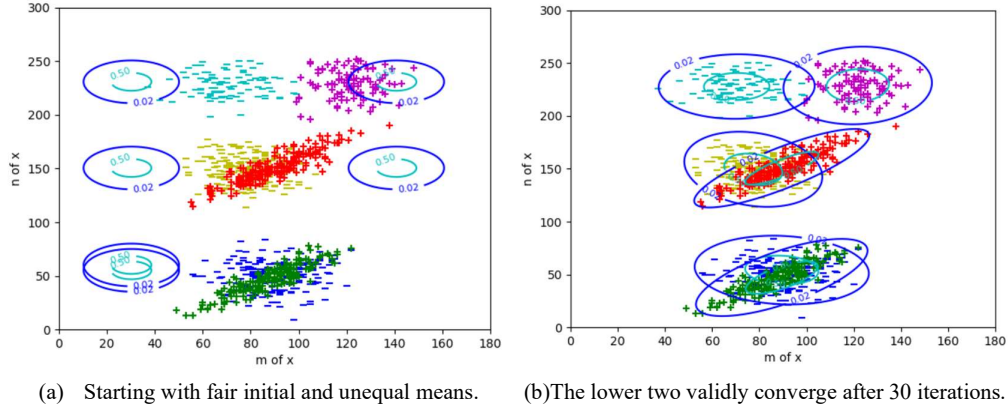

(a) Starting with fair initial and unequal means.  (b)The lower two validly converge after 30 iterations.

**Fig. 8**. Example 5 with better initial means for validly converge.

The author still used the bad initial parameters, as shown in Fig. 7(a), but with a larger sample whose size was 50000. In this case, the iteration could also validly converge, as shown in Fig. 9.

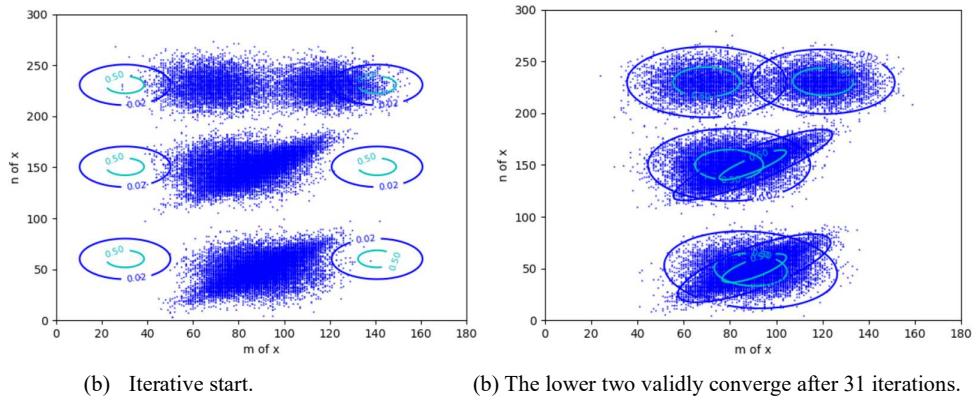

(b) Iterative start.  (b) The lower two validly converge after 31 iterations.

**Fig. 9**. Example 3 with bad initial parameters and a huge sample.

This example indicates that even if the overlap is severe, the CM-EM algorithm can also globally converge if the sample is big enough, or the initial means are fair and unequal. However, for some improperly initial parameters and smaller samples, the convergence may be invalid.

## 5. Discussions

### 5.1. About how results support conclusions

The author claims Affirmation I and Affirmation II (see Section 1) are two theoretical mistakes in the popular convergence theory of the EM algorithm for mixture models; we can improve the EM algorithm by the semantic information methods for better convergence.

Example 1 in Section 2.2 and Example 2 in Section 3.1 reveal that Affirmation I is wrong because the observed data log-likelihood $L$ and the complete data log-likelihood $Q$ are not always positively correlated. Affirmation II is also wrong because $Q$ may decrease and should decrease in some cases. Section 2.3 provides the mathematical analysis that explains why the global maximum of $Q$ is not $Q^*$ that results in the global maximum of $L$. Figure 5 in Sections 3.1 and Section 4.2 indicates that local convergence happens not because of locally maximal $Q$ but because of unfair competition and small samples.

The CM-EM algorithm is provided as an improved EM algorithm. Fig. 6 and Table 1 in Section 4.1 compare two algorithms. The results indicate that the CM-EM algorithm can save 26% of iterations that the EM algorithm needs. We can also accelerate convergence using the fair marriage principle. Example 3 shows that the CM-EM algorithm can also work well for a two-dimensional Gaussian mixture. As for the running time of each iteration, the CM-EM algorithm only requires a little longer time than the EM algorithm because the E2-step is simple, and we may repeat the E2-step no more than three times.

Fig. 6 with two groups of iteration numbers also provides a method for the comprehensive comparison between an improved EM algorithm and the EM algorithm.

### *5.2. About the boundary convergence and the limitation of the CM-EM algorithm*

Regarding how to avoid the local convergence and the boundary convergence, there have been many effective methods, such as the SMEM algorithm [5], the CEM algorithm [6], Random swap EM algorithm [7], and the Cross-entropy algorithm [8]. In these algorithms, exterior circulations are used to improve initial parameters or minor parameters. In the Random Swap EM algorithm, the method is to remove a component and add a component randomly. In the SMEM algorithm, the method is to split two components and merge two components. In the CEM algorithm, selecting split and merge operations is based on the competitive mechanism.

Using the CM-EM algorithm and the fair marriage principle introduced in Sections 3.2, we can easily avoid the local convergence and accelerate the global convergence of the binary Gaussian mixture. However, for the mixture of more components or samples in two-dimensional or multi-dimensional instance spaces, the fair marriage principle now is not sufficient. It is still a difficult task to enrich this principle and to convert it to practical methods. Nevertheless, the CM-EM algorithm does not exclude other improvements to the EM algorithm. How do we combine the existing algorithms with the CM-EM algorithm and the fair marriage principle for better convergence? We need further study.

### 6. Conclusions

The popular convergence theory of the EM algorithm for mixture models affirms that (1) the complete data log-likelihood $Q$ and the observed incomplete data log-likelihood $L(\mathbf{X}|\theta)$ are positively correlated, and (2) $Q$ is increasing in the M-step and non-decreasing in the E-step. Under the guidance of this theory, Ueda and Nakano proposed the DAEM algorithm, in which the annealing operation was used to avoid local convergence because of locally maximal $Q$. This paper used two examples, one of which was proposed by Ueda and Nakano, to show that the two affirmations in the popular convergence proof are wrong. By analyzing the example proposed by Ueda and Nakano, this paper concluded that local convergence exists not because there is locally maximal $Q$, but because the sample is too small, and the unfair competition exists.

The CM-EM algorithm almost retains the simplicity and the high-efficiency feature of the EM algorithm. On average, it can save 26% of iterations that the EM algorithm needs. However, for the mixtures of more components in multi-dimensional instance spaces, the CM-EM algorithm is still raw. It is expected that the CM-EM algorithm, together with the new convergence theory [16] and the fair marriage principle, could provide a solid foundation for other improvements to algorithms for mixture models.

This paper proposed the fair marriage principle to avoid local convergence. Using the initialization map, we should be able to save 90% of average iterations for binary Gaussian mixture models in one-dimensional instance space. However, for the mixtures of more components in two-dimensional or multi-dimensional instance spaces, we need further studies to enrich this principle and to convert this principle into specific methods.

*Appendix I. Supplemental Materials*

The supplemental materials including several Python 3.6 source codes for Examples 1-3 can be downloaded from http://survivor99.com/lcg/cm/Python4Ex1-4.zip

*Appendix II: The Proof of* $\frac{dH(Y|X,\theta)}{d\sigma} > 0$ *for Example 1.*

Letting $t = P(y_1|x)$, we have

$$\frac{dH(Y|X,\theta)}{dt} = \sum_i P(x_i) \frac{d}{dt}[-t\log t - (1-t)\log(1-t)]$$

$$= \sum_i P(x_i) \log \frac{1-t}{t} = \sum_i P(x_i) \log(e^{-b/\sigma^2}) = -\sum_i P(x_i) b/\sigma^2.$$

Since

$$\frac{dt}{d\sigma} = \frac{d}{dt}\left(\frac{1}{1+e^{-b/\sigma^2}}\right) = -\frac{2be^{-b/\sigma^2}\sigma^{-3}}{(1+e^{-b/\sigma^2})^2},$$

we have

$$\frac{dH(Y|X,\theta)}{d\sigma} = \frac{dH(Y|X,\theta)}{dt}\frac{dt}{d\sigma} = \sum_i P(x_i) \frac{2b^2 e^{-b/\sigma^2}\sigma^{-5}}{(1+e^{-b/\sigma^2})^2} > 0.$$

**QED.**